\documentstyle[harvard]{nemlap}

                           {\end{enumerate}}
                        {\end{it}\end{trivlist}}
                        {\end{itemize}\end{sl}}
\newenvironment{citaat}{\begin{quote}\small\sl}{\normalsize\end{quote}}
\newenvironment{opsomming}{\begin{itemize}\itemsep -0.7ex}{\end{itemize}}

\newcounter{ex}

\begin{document}
\title{A Machine Learning Approach to the Classification of Dialogue Utterances \\}
\author{Toine Andernach}
\institute{Parlevink Group, Department of Computer Science \\
University of Twente, P.O. Box 217, 7500 AE Enschede, The Netherlands \\
{\tt andernac@cs.utwente.nl}
}

\maketitle

\begin{abstract}
\sl The purpose of this paper is to present a method for {\em automatic}
classification of dialogue utterances and the results of applying that
method to a corpus. {\em Superficial} features of a set of training
utterances (which we will call {\em cues}) are taken as the basis for finding
 relevant utterance classes and for extracting rules for assigning these 
classes to new utterances. Each cue is assumed to partially contribute
to the communicative function of an utterance.
Instead of relying on subjective judgments for the tasks of finding classes
and rules, we opt for using {\em machine learning} techniques to guarantee objectivity. 
\end{abstract}

\section{Introduction}
\label{intro}


In tasks such as determining a taxonomy of dialogue acts or a set of
rules for assigning dialogue acts to utterances, researchers often rely on
subjective judgments or on classifications already available in
literature \cite{Hinkelman:90}, \cite{SeligmanFais:94},
\cite{LitmanPassonneau:95}, \cite{Carletta:96}.

In this paper, we will present an alternative approach to dialogue act
classification in which as many classification decisions as possible
are taken automatically, i.e. without intervention of a human being.
The advantages of this approach over more traditional approaches are
the consequent behavior of the machine and the ability to combine
different knowledge sources much faster than people can.  Our approach
differs from superficial approaches like \cite{Hinkelman:90},
\cite{SeligmanFais:94} and \citeasnoun{SchmitzQuantz:95} in that the
the set of dialogue acts and dialogue act rules are automatically
learned from a training corpus by machine learning techniques.

In section \ref{context} we will first give a brief description of the 
context of the research, the {\sc Schisma} project. 
In section \ref{cues} and \ref{schismacues} we will discuss the role
of surface linguistic information relevant for
dialogue act classification in general and in {\sc Schisma} respectively. 
Section \ref{schismacues} also presents {\em cue patterns} as a means for 
objectively representing surface linguistic information.

In our approach, cue patterns are the input for an {\em unsupervised} 
classification algorithm 
which outputs classes of cue patterns and which is discussed in section \ref{unsupervised}.
These classes, together with the cue patterns are input for a {\em supervised}
classification algorithm discussed in section \ref{supervised}.
This algorithm derives rules which describe the classification yielded
by unsupervised learning. The number and complexity of the rules can
be used as a measure of the quality of the classification yielded by
unsupervised learning.
Sections \ref{conclusions} and \ref{future} contain conclusions and
future work respectively.

\section{The context: the {\sc Schisma} project}
\label{context}

In {\sc Schisma} we aim at developing a natural language keyboard dialogue system 
which interfaces a database containing information about theater performances. 
The interface should allow people to enquire about 
performances in general, to tune in to a specific performance and, if
desired, make a reservation for this performance. Research until now
has concentrated on various aspects of realizing such a theater
information and booking system such as the building
of a Wizard of Oz environment for the acquisition of a corpus of
dialogues for this domain, analysis and tagging of the dialogue
corpus, recognition of domain-specific concepts, syntactic analysis
and dialogue modeling. The
utterances used as examples in this paper are all taken from the
corpus of simulated man-machine dialogues yielded by the Wizard of OZ experiments.

In {\sc Schisma}, we consider a {\em dialogue} to be a 
sequence of turns of two speakers. The word {\em Speaker} is used in the 
general
sense of the word: a language user. A {\em turn} is an uninterrupted stream
of language by one speaker. A turn consists of {\em utterances}: linguistically
identifiable units. Typically, typed or written dialogues utterances are
divided by punctuation marks and conjunctives.
A {\em dialogue act} is a communicative function expressed by a 
dialogue utterance.

\section{The role of surface elements: cue patterns}
\label{cues}

Many of our insights are based on Conversation Analysis  (CA)
\cite{SacksSchegloff:78}.  Conversational analysis is rule-governed
and the underlying idea is that shared knowledge of these rules (or
{\em conventions}) most often enables conversants to have smooth
flowing and coherent conversations with one another.



Like CA we assume that there is a strong interdependence
between speakers' wishes and the way they chose their utterances, i.e. 
between {\em form} and {\em function} of utterances in a dialogue.
The more we can rely on 
superficial information to exploit this rather complex interdependence
(see \cite{HinkelmanAllen:89}), the more computationally attractive this will 
be.

A surface-linguistic way of determining dialogue acts has already been 
suggested by several researchers.
\citeasnoun{HinkelmanAllen:89} use patterns of linguistic features to suggest
a range of speech act interpretations for an utterance. The dialogue act
classification rules they formulated, express {\em linguistic conventions}.

\citeasnoun{SeligmanFais:94} describe an approach to speech act interpretation
in the context of a speech translation system; they restrict their attention
to communicative goals which can be explicitly expressed via so-called conventional 
surface {\em cue patterns} rather than deep intentions. 
We will use the following definitions of the terms {\em cue
  patterns}\/, {\em cue}\/ and {\em cue value}\/ (based on \cite{SeligmanFais:94}:

\begin{description}
\item[cue pattern] a configuration (tuple) of one or more cue-cue value pairs
\item[cue] any aspect of the surface syntax or morphology of a language
\item[cue value] an instantiation of a cue
\end{description}

{\em Cues}\/ must not be confused with {\em cue words}\/; The latter term will
be used for special words which contribute to the communicative function
of an utterance (see section \ref{schismacues}). 
We will adopt this way of representing linguistic knowledge by cue
patterns because these structures are very suitable for our purpose of
automatically classifying dialogue utterances.




\section{Describing the data: cues in {\sc Schisma}}
\label{schismacues}

In this section we will discuss the utterance {\em cues}\/ which are
used as the empirical basis for dialogue act classification in {\sc Schisma}.

The initial set of cues used for dialogue act interpretation is  given
in table \ref{cuevalues}. 

\small
\begin{table}[htb]
\center
\begin{tabular}{|lr||l|} 
\hline
Cue & Label & Description\\
\hline
\hline
Utterance Type &  (UT)   & the {\em mood}\/ of an utterance or the syntactic category in case of a phrase \\
Wh-word?  & (WH)        & the presence or absence of a wh-word \\
Subject Type &  (ST) & the type of the subject of an utterance (if present)\\
Cue Words &      (CW)   & words which change the communicative function (if present)\\
First Verb Type & (FVT)   &  the type of the first verb in the utterance (if present)\\
Second Verb Type & (SVT)   &  the type of the second verb in the utterance (if present) \\
Question mark? & (QM)   &  the presence or absence of a question mark\\
\hline
\end{tabular}
\caption{\label{cuevalues}Cues and their potential values}
\end{table}
\normalsize

This set of cues is based on cues used in
literature ({\em Utterance Type, Cue Word}) and on intuitions about
their usefulness; {\em Subject Type}, {\em First Verb Type} and {\em
 Second Verb Type} are selected because of their potential
informativity on kinds of agents and actions. The optimal set of cues and
cue values however, can not a priori be selected; they will be the
result of the iterative process of cue set construction, tagging,
testing, interpretation and updating (see section \ref{automatic}).

Furthermore, the extraction of cue patterns from utterances is not
done automatically; a parser which could identify the cue pattern of all
utterances in our corpus is not yet available. We are, however, 
writing a grammar for the {\sc Schisma} domain and use the left-corner
parser for PATR-II developed at SIL (PCPATR) \cite{Shieber:86}.
Recent uses of the parser showed that no principle problems need  be expected
in automating cue pattern extraction; the set of cues and their
potential values concern objectively observable linguistic elements.

(\ref{Antigone}) is taken from the corpus and is represented by the
{\em cue pattern} in table \ref{cuepattern}:
\begin{example}
{\sl ik wil graag meer weten over de voorstelling 'Antigone'.}
\label{Antigone}
\end{example}

\begin{table}[htb]
\center
\begin{tabular}{|c|c|c|c|c|c|c|} 
\hline
UT & WH  & ST       & CW    & FVT & SVT & QM \\
\hline
\hline
DEC  & -   &  i   & graag & w  & e & -  \\
\hline
\end{tabular}
\caption{\label{cuepattern}A sample cue pattern}
\end{table}

The cue pattern in table \ref{cuepattern} expresses that
(\ref{Antigone}) has a declarative utterance type, does not contain a
wh-word nor a question mark, has a first person pronoun as subject, {\em willen}\/ as first verb type, a cognitive verb as second verb
type and contains the cue word {\em graag}.

\subsection{Utterance type}

The first cue we discuss is {\em utterance type (UT)}.
The corpus contains utterances with an utterance type which more or less 
corresponds to {\em sentence mood}\/ and utterances which do not.

Table \ref{utterancetypes} gives an overview and a description of the utterance types in
{\sc Schisma}.

\begin{table*}[htb]
\center
\begin{tabular}{|c|l|} 
\hline
Label & Description \\
\hline
\hline
DEC &   finite verb in second sentence position and no wh-word \\
WHQ &   wh-word in initial position  \\
YNQ &   a finite verb in initial position and a subject in second position \\
IMP &   a 1st pers. sing. or 3rd pers. sing. verb in initial position \\
PRE &   (one or more) prepositional phrases \\
NOM &   (one or more) nouns, noun phrases or proper names \\
ADJ &   (one or more) adjectives, adverbs or numbers \\
THA &   thank \\
GRE &   greet \\
CON &   confirmations, negations \\
EXC &   interjections, emotives, exclamations \\
MIS &   miscellaneous \\
\hline
\end{tabular}
\caption{\label{utterancetypes}Utterance Types}
\end{table*}


Many utterances in the corpus are regular {\em declarative sentences (DEC)}.
They are defined as having a finite verb in second sentence position
and no wh-word in (a constituent in) sentence initial position.
An example from the corpus is (\ref{d1}):

\begin{example}
{\sl Ik wil graag een reservering.}
\label{d1}
\end{example}


{\em YNQ} utterance types are assigned to utterances with a finite 
verb in sentence initial position and a subject in second position.


All utterances which contain a wh-word in (a constituent in) initial
sentence position are considered to be of type {\em WHQ}.
{\em IMP}\/ utterance type is assigned to utterances with a first person
singular or third person singular verb in sentence initial position.
Furthermore, the utterance should not contain a subject.
Another instance of {\em IMP}\/ utterance type concerns utterances with
an infinite verb in sentence-initial position. The {\em IMP}\/
utterance type roughly corresponds to the traditional imperative
sentence mood.

Utterances types which do not correspond to one of the conventional
sentence moods as described above typically consist of a (sequences
of) phrases (multiple PPs or multiple 
NPs), polite thanking and greeting utterances and typically short forms like
confirmations, negations, interjections and exclamations.

(\ref{p}) is an example of utterance type {\em PRE} from the corpus:

\begin{example}
{\sl En op 18 maart?}
\label{p}
\end{example}

Note that although (\ref{p}) is not a strict PP,
we consider it to be one; conjunctives like {\em en}\/ and {\em of}\/ are 
considered to be absent for the task of determining utterance types.
This also applies to conventional utterance types.
While (\ref{p}) only contains a rather atomic PP, there are also cases in
which the PP has a complex structure.
Utterance type {\em NOM} is assigned to (sequences of) Ns and/or NPs.
Most utterances tagged with this utterance type consist of one noun.
They function either as answer to former questions or as prompts to provide
all instances of a certain domain type but complex
NPs do also occur.
A collection of ADJ(P)s, ADV(P)s and Numericals is represented in the
{\em ADJ}\/ type. Again, most of the occurrences are one word
utterances, consisting of a number or an attitude adverb.


The kind of utterances belonging to the  {\em CON}\/ type (negation
markers also belong to this class), {\em THA}\/ type, {\em EXC}\/ type
(which also contains exclamatives and  interjections) and {\em GRE}\/
type should be obvious. The utterance type {\em MIS}\/ functions as a
kind of a garbage collector; all utterances which cannot be classified
as one of the utterance types above are covered by this type.








\subsection{Other cues}

Another cue for the communicative function of
utterances is the presence of a {\em wh-word (WH)}. This cue might
seem to be superfluous because of the presence of the WHQ utterance
type. However, wh-words do not necessarily occur in utterances with a
wh-question type. (\ref{stuk}) has a YNQ Utterance Type and contains
the wh-word {\sl wie}.

\begin{example}
{\sl WEet u wie dat stuk heeft geschreven?}
\label{stuk}
\end{example}


Like WH, The use of {\em Question Mark (QM)}\/ as potential cue is not
as trivial as
it might seem to be. After a closer look at the corpus we found that 
utterances with a {\em WHQ}\/ or a {\em YNQ}\/ utterance type are not always
ended with a question mark as we might expect. On the other hand,
many utterances with utterance type {\em DEC}\/ do have a question mark: 

\begin{example}
{\sl 2 zei ik toch?}
\label{toch}
\end{example}

See \cite{Beun:89} for a discussion of so-called {\em declarative questions}\/
like (\ref{toch}).


The cue {\em Subject Type (ST)} is only applicable to utterances with
utterance types {\em WHQ}\/, {\em YNQ}\/ and {\em DEC}.
We distinguished the following superficial subject types:

\begin{opsomming}
\item NPs referring to artists, performances, tickets, discounts etc. like
{\em Herman Finkers}
\item First person personal pronouns like {\em ik}\/ referring to the speaker 
\item Second person personal pronouns like {\em u}\/ referring to the hearer
\item Third person personal pronouns like {\em het}
\item Interrogative pronouns like {\em welke}\/, {\em wanneer}\/ etc.
\item Demonstrative pronouns like {\em dit}\/ and {\em dat} 
\item Others
\end{opsomming}


Another kind of cues for the function of utterances in discourses are 
{\em Cue Words (CW)}. Most cue words are realized as modal adverbs or adverbial
phrases. {\em graag} for example, can be translated as {\em like to}.
In dialogue however, it is often used as a more general politeness 
marker like in (\ref{graag}):

\begin{example}
{\sl Ik wil graag naar {\em Mini en Maxi}.}
\label{graag}
\end{example}

While used in a declarative utterance, {\em graag}\/ intuitively
strengthens the wish for information or action.
(\ref{graag}) shows that cue words can be very 
subtle indications for speaker intentions in discourse, very often in 
combination with other cues in utterances.





The last cues are {\em First and Second Verb Type (FVT/SVT)}\/. The
corpus contains both utterances with and without verbs. 
After analyzing the corpus, we found the following main kinds of verbs:

\begin{opsomming}
\item verbs like {\em to reserve}, {\em to be}\/, {\em to have}\/ and {\em to know}
\item domain verbs, which only have domain objects as their arguments
\item task verbs, which have at least one argument which refers to a 
dialogue partner
\item auxiliary verbs
\item not applicable
\end{opsomming}

Many utterances contain more than one verb and most do not contain
more than two. Therefore, we assume that the types of the first two
verbs are the most relevant ones.

In the next sections we will discuss our way of automatically
classifying dialogue utterances.

\section{Towards the automatic classification of dialogue utterances}
\label{automatic}

The central idea of our approach to dialogue act classification is to
automatically derive dialogue acts and dialogue act rules from a corpus.
We would like to prevent human subjective judgments as much as
possible because people are not very good at deriving rules from a
rather large set of examples, like a corpus of dialogue utterances.
\citeasnoun{LitmanPassonneau:95} for instance, found in an 
experiment that performance of individual speakers varied widely when they 
were asked to indicate segment boundaries in discourses, a task
comparable to the one at hand.

We also experienced these problems in our attempts to tag the {\sc
  Schisma} corpus with dialogue act labels; it appeared to be very
difficult to consequently determine the dialogue act of an utterance because
there were no strict criteria available to decide {\em when} to assign 
{\em which} dialogue act label.
This complexity of dialogue act classification (also noted 
by \citeasnoun{HinkelmanAllen:89}) lends us more support for our choice to
use algorithms implemented in computer programs instead.
Furthermore, a machine can do the job much faster.



The term {\em classification} is often used in two senses: it either means 
assigning a class (element of an existing set of classes) to a new case, or 
finding the classes themselves from a given set of unclassified cases.
The latter is often called {\em unsupervised classification} as opposed to
the former {\em supervised classification} in which the set of classes must be
 provided beforehand.
In this section we will propose a way of using both kinds of classification
in determining the dialogue act of a given utterance.



After having tagged the whole dialogue corpus  yielded by the Wizard of
Oz experiment (64 dialogues) with relevant cues already
discussed in section \ref{schismacues} the corpus was converted to the
generic mark-up language SGML. This converted corpus served as the
input for scripts to create sub-corpora for the different kinds of
algorithms used. 

First, the utterances in the corpus (2351) were randomly split in a
training set which consisted of 75\% (i.e. 1763) of all utterances and
a test set of 25\% (i.e. 588) of all utterances.

After that, an {\em unsupervised} classification algorithm was applied
to the training set to automatically discover dialogue acts (see
section \ref{unsupervised}). The output consists of a list of
all cue patterns in the training set together with their class (see section
\ref{supervised}).

Then, we applied a {\em supervised} classification algorithm to the output of
the unsupervised classification algorithm to extract a reduced
dialogue act rule set.  The rules yielded by the supervised classification
algorithm had the form of if-then rules.
In the following sections, we will discuss the classification process in the 
training phase and the testing phase in more detail.

\subsection{Unsupervised classification}
\label{unsupervised}
 
Unsupervised classification (also called {\em clustering}) is concerned with
the automatic discovery of classes in data. Classification is based
on the fact that some cases are more like each other than the rest of
the cases. 

\subsubsection{Methodology}
\label{unsupmethodology}

We used the program {\em AutoClass} \cite{CheesemanStutz:95}
in which a method for unsupervised learning is implemented. 
 AutoClass is an unsupervised Bayesian classification system that seeks a 
maximum posterior probability classification.
In the Autoclass approach, class membership is expressed
probabilistically; every item is considered to have a probability that it
belongs to each of the possible classes.

AutoClass allowed us to automatically find the set of classes that is
maximally probable with respect to cue patterns in the training set. 
These classes should resemble classes which we would indicate as
dialogue acts.  

However, \citeasnoun{CheesemanStutz:95} stress the fact that the
discovery of important classes in data is rarely a one shot process. Instead,
it is:
\begin{citaat}
`` (...) a process of finding classes, interpreting the results, transforming
and/or augmenting the data, and repeating the cycle.''\\
(Cheeseman and Stutz 1995:62)
\end{citaat}

This implies the intervention of an expert to judge the intermediate
results and to transform or augment the data if necessary; the number
or kind of cues, the number of AutoClass trials or some other
parameters could be changed. 
\citeasnoun{CheesemanStutz:95} also stress the need for a strong interaction
between the classification program and the expert, because their contributions
to the classification process are complementary; the program can handle a
huge amount of data with a high speed, while the expert has domain knowledge.

Instead of the intervention of a human being however, we could also
look for ways
to automatically judge the appropriateness of the set of classes, for 
instance by taking the highest common factor of the results of multiply
applying Autoclass to different (randomly generated) training sets.

AutoClass Input consisted of the 75\% training set of cue patterns. The cues
chosen for the first experiment were {\em Speaker}, {\em
  UtteranceType}, {\em SubjectType}, {\em FirstVerbType} and {\em
  QuestionMark}. 
AutoClass was instructed to report on the best two classifications found.
AutoClass reports contained information about the strength of the
classes found and the relative importance of the cues for a specific class.
Furthermore, both case and class descriptions were given for every cue pattern
in a certain class.
The results of the first runs of AutoClass on our training corpus will
be discussed in the following section.

\subsubsection{Results}
\label{unsupresults}

We let AutoClass generate five classifications of which the two best
were stored. These two yielded ten and seven classes respectively, for 206 
different cue patterns in the training set, i.e. about 20 and 30 cue
patterns in each class.

We will now focus on the classification which yielded seven classes
and informally discuss the kind of cue patterns in every class.
Class 0 can be roughly identified as consisting of cue patterns for which
SubjectType='n' (no subject present) or UtteranceType='n' (Nouns and
NPs). In the former respect this class is similar to Class 6.
Class 0 and Class 6 utterances together can be informally
characterized as all {\em utterances without a subject}.

Class 1 only consists of UtteranceType='d' patterns (declarative
utterances) and does both contain Client and System utterances.
Other patterns in this class typically have  SubjectType='a' or 'p' (artist or performance) or FirstVerbType='z' or 'd' (a form of {\em to be} or a domain verb). 
The features of Class 1 can roughly be characterized as {\em Information supplying utterances about domain objects}.

Class 2 cue patterns have either UtteranceType='w' or 'y'
(wh-utterances and yes/no utterances) and SubjectType='i' or 'e'
(first person or second person pronouns). A suitable common description
for the cases in Class 2 is {\em questions concerning actions of one
of the conversation participants or states in which they are}.

Class 3 patterns only contain Client utterances with UtteranceType='w'
or 'y'. In this latter sense, they are related to Class 2.
The main distinction however, is that Class 3 patterns all concern
{\em  Client questions about facts in the domain}. Note that System
questions about domain facts do not occur in the corpus.

Class 4 utterances are tagged with UtteranceType='d'. Most of them
also have SubjectType='e' (second person pronoun), which means that most of the Class 4
utterances are {Information supplying utterances about actions of {\em
    the other} conversation participant or the state where he is in}.
Strange enough, we also found some patterns in Class 4 which at first sight, are
not similar to the other cases.

Most of the Class 5 utterances are both UtteranceType='d' and
SubjectType='i' (first person pronoun) utterances. Some others have
SubjectType='t' (demonstrative pronoun) instead of 'i'. Most Class 5 Client
utterances have FirstVerbType='w'. These are the utterances in which
Clients express a wish.

Table \ref{classstrength} shows the relative strength of the
classes found. This measure is based on the mean probability of instances
belonging to each class and provides a heuristic measure of how
strongly each class predicts its instances.

\begin{table}[htb]
\center
\begin{tabular}{|c||c|} \hline
Class & Relative class strength \\ \hline\hline
0     & 0.596 \\ \hline
1     & 0.705 \\ \hline
2     & 0.167 \\ \hline
3     & 0.067 \\ \hline
4     & 1.000 \\ \hline
5     & 0.331 \\ \hline
6     & 0.199 \\ \hline
\end{tabular}
\caption{\label{classstrength}Relative class strength}
\end{table}

It is clear from table \ref{classstrength} that classes 4 and 1 are
the strongest and class 3 is the weakest class.
Together with the observations of the classified data, the
information given in table \ref{classstrength} can be used in the
process of deciding which classes are going to be used in the eventual
system.

Table \ref{atribimp} shows the relative influence of each cue in
differentiating the classes from the overall set of cue patterns.

\begin{table}[htb]
\center
\begin{tabular}{|c||c|} \hline
Cue   & Relative influence \\ \hline\hline
Sp    & 0.270 \\ \hline
Ut    & 0.859 \\ \hline
St    & 1.000 \\ \hline
Fvt   & 0.881 \\ \hline
Qm    & 0.248 \\ \hline
\end{tabular}
\caption{\label{atribimp}Relative influence of cues}
\end{table}

The cues Question Mark and Speaker appear to have little
influence compared to the other three cues whose influence is about
three to four times as great. This information can be used for new
trials of AutoClass with fewer cues, omitting cues with little
influence. Information about the influence of the {\em values} of each cue is
also generated by AutoClass. This can also be used for new trials with
other cue values.

In the next section we will discuss {\em supervised classification} and
  the way we applied it to our dialogue corpus.


\subsection{Supervised classification}
\label{supervised}

To gain more insights in the results of unsupervised classification,
supervised classification has been applied to the output of
the unsupervised classification algorithm yielding a reduced
dialogue act rule set.
A set of cue patterns with their classes can be regarded as a set of
rules in which the number of cue patterns equals the number of rules.
The goal of supervised classification is then, to reduce the number of
rules i.e. to extract generalizations from the data. Such a reduced
rule set should allow us to assign classes to new cue patterns.

\subsubsection{Methodology}




We used the supervised classification program {\em CN2}
\cite{ClarkBoswell:91} which induces if-then rules. 
It is designed to work well for domains where there might be noise, which is 
the case in our domain; instead of using normal significance testing or
entropy which favor rules which cover examples of only one class, it uses
the so-called Laplace accuracy which does not have a downward bias; rules
which cover many examples of one class and a few examples of another class
are preferred to rules which cover a few examples of only one class.

We applied CN2 to the same training set of cue patterns as we used for
discovering classes by unsupervised learning. 
CN2's application in {\sc Schisma} takes as first input the fixed set
of cues and their potential values and the set of classes (dialogue acts) to 
be learned. The second input consists of the training data, i.e. a set of cue
patterns together with their class yielded by the unsupervised
classification phase. The output of CN2 is a set of if-then rules, which 
predict classes given one or more cue--cue value expressions.

We consider the number of rules relative to the number of different
cue patterns as one indication of the quality of the rule set. 
More specifically, the lower this number the more general the rule
set. We call this metric {\em the specificity index (SI)}:
\bigskip

$ SI = \frac{\# rules - (\# classes-1)}{\# cue pattern types - (\# classes-1)}$
\bigskip

The underlying idea of this index is that maximum abstractness
is obtained if SI approaches zero, i.e. if few rules describe many CPTs. 
SI can be used to compare the abstractness of different rule sets.

\subsubsection{Results}

Applying CN2 resulted in a set of 44 rules for 206 different cue patterns
types (CPTs) with an SI of 0.19. Two of the rules are shown in figure \ref{cn2rules}
(cue names and value names are changed to improve readability):

\begin{figure}[htb]
\begin{verbatim}
IF    SUBJECTTYPE = 2nd person pronoun
  AND QUESTIONMARK = yes
THEN  CLASS = 2  [0 0 96 0 0 0 0]

IF    UTTERANCETYPE = Noun/NounPhrase
THEN  CLASS = 0  [109 0 0 0 0 0 0]
\end{verbatim}
\caption{\label{cn2rules}CN2 Rules}
\end{figure}

The first rule informally expresses that if an utterance contains a
subject which is a second person pronoun and a question mark, than it
will be assigned class 2, i.e. it is a question concerning an action of one
of the conversation participants or state which he is in (see
section \ref{unsupresults}). The second rule expresses that all
utterances which only consist of a noun or noun phrase are assigned
class 0. 

Due to the fact that we chose to generate an unordered rule set
instead of an ordered one for reasons of interpretability,  a partial
cue pattern could give rise to more than one class. Suppose that the
rules in figure \ref{cn2rules} were the only ones and that an unknown cue pattern
  matched with both of them. In that case, the distributions would be
  summed and frequencies calculated; Class 2 would have a probability
  of 96/205 and Class 0 of 109/205 which means that Class 0 would be
  assigned to the unknown cue pattern, because of its higher probability.

Table \ref{eval} expresses the accuracy with which the rules generated
from the training set describe the unsupervised classification of the
test set. 
\begin{table}[htb]
\center
\begin{tabular}{|c||c|c|c|c|c|c|c|c|} \hline
Actual$\backslash$Predicted    &   0  &   1   &  2  &   3  &   4  &   5 &    6 &   Accuracy \\ \hline\hline
0 &  146  &    0    &   0   &    1   &    0   &    0  &     0   &   99.3 \% \\ \hline
1 &    3  &   131   &    0  &     0  &     0  &     0 &      0  &    97.8 \% \\ \hline
2 &    0  &     0   &   75  &     0  &     0  &     0 &      0  &   100.0 \% \\ \hline
3 &    2  &     0   &    0  &    66  &     0  &     1 &      0  &    95.7 \% \\ \hline
4 &    6  &     0   &    0  &     0  &    78  &     2 &      0  &    90.7 \% \\ \hline
5 &    1  &     0   &    0  &     0  &     0  &    43 &      0  &    97.7 \% \\ \hline
6 &    0  &     0   &    0  &     0  &     0  &     0 &     33 & 100.0 \% \\ \hline
\end{tabular}
\caption{\label{eval}Evaluation results of the set of supervised
  classification rules}
\end{table}

The accuracy of the rules for a certain class {\em c} can be expressed by the
ratio of the number of well-predicted cue patterns and the total
number of cue patterns in {\em c}. This metric is also called {\em
  precision}.
Table \ref{eval} shows the number of cue patterns for every possible
pair of actual and predicted classes. The accuracy as defined above is
included in the last column.

\section{Conclusions}
\label{conclusions}

As was already noted by \citeasnoun{CheesemanStutz:95} (see section
\ref{unsupmethodology}) and as we experienced ourselves,
our way of automatically finding dialogue utterance classes from a
corpus of training utterances {\em is not a one shot process}:
it requires iteratively training and testing with varying sets of cues
and cue values. In this process, the combination of domain knowledge
of the expert and the objectivity and computational power of the
machine is necessary. 

The classification results described however, are promising; the
classes yielded by the unsupervised classification algorithm can be
interpreted intuitively; for the task of predicting client and system
actions, however, the classes might be too general. More specific
classes will be obtained if the number of initial classes for
AutoClass is increased.

The information about the relative influence of cues
and classes generated by AutoClass is very useful for the training-testing cycle.
The problem of finding the optimal set of cues and cue values to train
the algorithms will also be solved during this cycle.

Furthermore, as shown by the evaluation results in table \ref{eval}
the supervised classification algorithm
found an acceptable general and accurate set of rules for the
 classes yielded by the unsupervised classification algorithm.

\section{Future Work}
\label{future}

At this moment, we are in the cycle of testing different sets of cues and their
values, generating and interpreting classes, generating rules for
deriving classes and re-testing on the basis of the results.
In parallel, a grammar is written which should allow us to automatically
extract cue patterns from the utterances in the training corpus.

In the near future, we will test alternative ways of supervised and 
unsupervised classification in {\sc  Schisma}. We will use the
unsupervised classification program package C4.5 \cite{Quinlan:93}
which can both yield rules and decision trees. It has some extra
features like n-folded cross-validation and the consultation of the
decision tree and rules.
Regarding unsupervised classification, we are currently working on the
generation of Kohonen Maps for discovering classes in the
training corpus. The results will be compared with the AutoClass results.

Another aspect paid attention to, is the use of
context information in the classification process. 
One could argue that at least some context information is necessary for finding
useful classes of dialogue utterances. We will test this hypothesis
by adding the class of cue pattern {\em n-1} as a cue to cue pattern
{\em n} for all cue patterns in the training set.
Another possible way of using local context information is applying
an n-gram analysis to the classes of all cue patterns in the training
set. To approximate the conditional probability $P$ that utterance
{\em n} falls in class {\em c} given the classes of m previous utterances deleted interpolation \cite{Jellinek:90}
will be used. This approach is similar to the one adopted  by
\citeasnoun{Reithinger:95} in the {\sc Verbmobil} project and has already been applied with promising
results. 





\bibliographystyle{agsm}




\end{document}